\documentclass[twocolumn]{bytedance_seed}

\usepackage[toc,page,header]{appendix}

\usepackage{enumitem}
\usepackage{microtype}
\usepackage{graphicx}
\usepackage{multirow}
\usepackage{booktabs}
\usepackage{amsmath}
\usepackage{wrapfig}
\usepackage{algorithm}
\usepackage{algpseudocode}

\usepackage{hyperref}

\usepackage{xcolor}

\usepackage{subcaption}
\usepackage{tcolorbox}
\usepackage{microtype}
\usepackage{mathtools}
\allowdisplaybreaks
\usepackage{soul}
\usepackage{subcaption}
\usepackage{graphicx}

\usepackage{siunitx}
\usepackage{CJKutf8}
\usepackage{xspace}
\usepackage{adjustbox}
\usepackage{amsmath, amssymb}

\usepackage{minitoc}

\newcommand{\name}{DisagMoE\xspace}

\title{DisagMoE: Computation-Communication overlapped MoE Training via Disaggregated AF-Pipe Parallelism}

\author[1, 2]{Zhichen Zeng}

\author[1, 3]{Chi-Chih Chang}
\author[2]{Jiayi Wang}
\author[2]{Zezhou Wang}

\author[1]{Ningxin Zheng}
\author[1]{Zheng Zhong}

\author[1]{Cesar A. Stuardo}
\author[1]{Dongyang Wang}
\author[3]{Mohamed S. Abdelfattah}
\author[1]{Haibin Lin}

\author[2]{Banghua Zhu}
\author[2]{Ang Li}
\author[1]{Ziheng Jiang}

\affiliation[1]{ByteDance Seed}
\affiliation[2]{University of Washington}

\affiliation[3]{Cornell University}

\abstract{
Mixture-of-experts (MoE) architectures enable trillion-parameter LLMs with sparsely activated experts.
Expert parallelism (EP) is a widely adopted MoE training strategy, but it suffers from severe all-to-all communication bottlenecks, which is exaggerated by the limited inter-node network bandwidth as the growing model size requires distributing experts across GPU nodes.
Prior work focused on overlapping these all-to-all communications with feed-forward network (FFN) and self-attention computations,
which often leaves residual network-bound stalls due to inherent imbalance in attention and FFN layers' computation-communication ratios.
We present \textsc{\name}, a disaggregated MoE training system that jointly optimizes model placement and scheduling for maximal efficiency.
\name separates attention and FFN layers into disjoint GPU groups, introduces a multi-stage pipeline with uni-directional, many-to-many communications, and employs a computation-communication roofline model to balance GPU and network bandwidth allocation among the attention and FFN groups.
\name is implemented on Megatron-LM,
and evaluation shows that \name improves training efficiency across multiple MoE models with up to 1.8$\times$ speedup on 16-node 8$\times$H800 clusters.
}

\date{\today}

\begin{document}
\maketitle

\section{Introduction}

Mixture-of-Experts (MoE) architectures
are widely adopted
for scaling Transformer-based large language models (LLMs)~\cite{shazeer2017outrageously, fedus2022switch, chowdhery2023palm, jiang2024mixtral}.
By partitioning the feed-forward network (FFN) layers into experts and activating only a small subset of experts for each token, MoE architectures decouple activated parameter count from total parameter count, achieving sublinear growth of per-token computation while scaling model capacity~\cite{lepikhin2020gshard, deepseek_v32, qwen_qwen36_max, zai_glm51, teamseed2}.
DeepSeek-V4-Pro, for example, scales to 1.6T total parameters while activating only 49B parameters per token~\cite{deepseek_v4_pro, dai2024deepseekmoe}.

\begin{figure}[!t]
    \centering
    \includegraphics[width=1.0\linewidth]{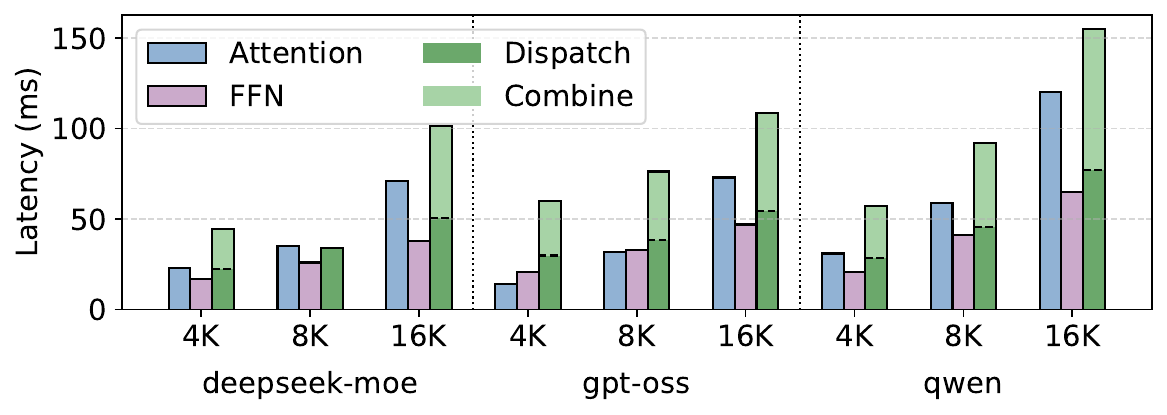}
    \caption{Analysis of the execution of MoE. Time breakdown of MoE models executed on 8 nodes H800 GPUs using Megatron-LM.}
    \label{fig:latency_breakdown}
\end{figure}

Despite the compute efficiency, large MoE layers still face GPU memory capacity challenges due to the total parameter count.
Expert parallelism (EP) addresses the challenge by sharding experts across GPU devices, where dense components such as attention layers are replicated via data parallelism (DP) among the same GPU group.
Two all-to-all communications arise from this EP+DP strategy, where tokens must be shuffled from the micro-batch ranking in attention layers to expert ranking in FFN layers (\textit{dispatch}) and then shuffled back (\textit{combine}).
These communications transmit almost the entire micro-batch due to dynamic expert routing, and they sit right on the critical path between the attention and MoE layers.
Moreover, as model size continues to scale, FFN layers no longer fit in a single multi-GPU node, forcing EP domain to span multiple nodes, where low inter-node bandwidth further exaggerates the communication bottleneck.
As shown by our profiling on 64$\times$\,H800 GPUs, the communications account for up to 50\% time of a training step (Fig.~\ref{fig:latency_breakdown}).

Optimizing inter-node dispatch/combine is a first-order optimization goal for large-scale MoE training, and a rich literature exists in this space~\cite{rasley2020deepspeed, rajbhandari2022deepspeed, he2022fastermoe, tutel, zhang2025comet}.
One line of work overlaps communication with computation.
Tutel \cite{tutel} and Comet \cite{zhang2025comet} propose operator-level chunking, which partitions FFN computations into fine-grained tiles and partially overlaps dispatch/combine with the MoE layer through tile-level pipelining.
However, as shown in Fig.~\ref{fig:motivationchunk}, the overlap window is bounded by FFN computation, leaving residual communication exposed.
Lancet \cite{jiang2024lancet} and DualPipe \cite{deepseekai2025deepseekv3technicalreport} enlarge the overlap window by pipelining multiple micro-batches and hiding the communications of one micro-batch with the computations of another micro-batch.
Nevertheless, they still face a fundamental imbalance:
as shown in Fig.~\ref{fig:motivationchunk}, attention computation time grows quadratically with sequence length, whereas FFN computation and all-to-all communications scale roughly linearly. This imbalance makes perfect overlap difficult across
a wide range of workloads, as shown in Fig.~\ref{fig:roofmotivation:a}.

Recent works~\cite{wu2025hetermoe, wang2025step, zhu2025megascale} disaggregate attention and FFN (AFD) onto separate device groups to overlap inter-group communication.
For serving, systems such as MegaScale-Infer \cite{zhu2025megascale} and StepFun \cite{wang2025step} place attention and FFN layers on heterogeneous hardware based on the observation that during decoding, attention layers are relatively more memory-bound and FFN layers more compute-bound. While effective for latency-driven serving \cite{wang2025step, zhu2025megascale}, this design is hard to apply to training: each group must hold its component's full parameters across all layers, plus training-only states (e.g., optimizer).
HeterMoE \cite{wu2025hetermoe} brings AFD to training onto heterogeneous hardware, and proposes zebra parallelism (ZP) between the FFN group using older GPUs and the attention group adopting newer GPUs, which are communicated by all-to-all primitives.
However, this two-stage pipeline is insufficient for large-scale MoE training, because each device group must retain all the parameters of their corresponding component across all transformer layers and causes out-of-memory at scale (Fig.~\ref{fig:expvpp}), with HeterMoE demonstrated only up to around 4.3B parameters, leaving how to scale to hundred-billion-parameter communication-overlapped MoE training an open question.
Moreover, existing methods maintain a fixed Compute-Communication ratio,
limiting the overlapping design space for balancing communication and compute across pipeline stages.

To address this imbalance and achieve scalable overlap training, we present \textsc{\name}, a large-scale MoE training system that disaggregates attention and FFN across device groups with adaptive worker allocation, jointly optimizing model placement and scheduling to mitigate all-to-all bottlenecks. Our key contributions are:

\textbf{Disaggregated component placement.}
\name{} partitions the model by \textit{component type}, assigning attention and FFN layers to different worker groups.
The attention groups replicate the dense components and distribute training data in micro-batches, while the FFN groups shard the experts and process tokens based on expert selection.
Each group may own components of the same type from multiple transformer layers,
thereby maximizing device utilization under GPU memory constraints.

\textbf{Multi-stage pipeline with overlapping.}
We propose \textit{AF-Pipe}, a multi-stage schedule that replaces the conventional all-to-all with inter-group \textit{many-to-many} (M2N) token exchanges and treats them as a first-class stage alongside attention and FFN compute, aligning stage boundaries across groups to systematically overlap communication with both computations.

\textbf{Adaptive worker allocation.} \name{} employs an adaptive allocator that \textit{jointly} sizes the attention and FFN groups as well as reallocating NIC bandwidth across them to balance the communication and computation time with various sequence lengths, top-$k$ selection criteria, and EP sizes.
Guided by a lightweight Compute-Communication roofline model, this resizing expands the effective overlap window, reduces critical-path communication, and mitigates the attention–FFN imbalance that limits prior overlap strategies.

We perform extensive pre-training experiments on MoE models with hundreds of billions of parameters, using clusters of 128$\times$H800 GPUs and sequence lengths ranging from 4K to 32K. \name{} consistently outperforms strong baselines, achieving up to 1.81× higher throughput than Megatron-1F1B interleaved overlap and up to 1.34× higher than state-of-the-art MoE overlap systems.
\section{Background and Motivation}

\subsection{Mixture of Experts (MoE) Structure}

The Mixture of Experts (MoE) architecture is a key technique for scaling LLMs efficiently while improving performance \cite{deepseekai2025deepseekv3technicalreport}. The MoE architecture substitutes a standard feed-forward network (FFN) with a set of multiple FFNs, known as \textit{experts}.

\begin{figure}[t]
    \centering
    \includegraphics[width=1.0\linewidth]{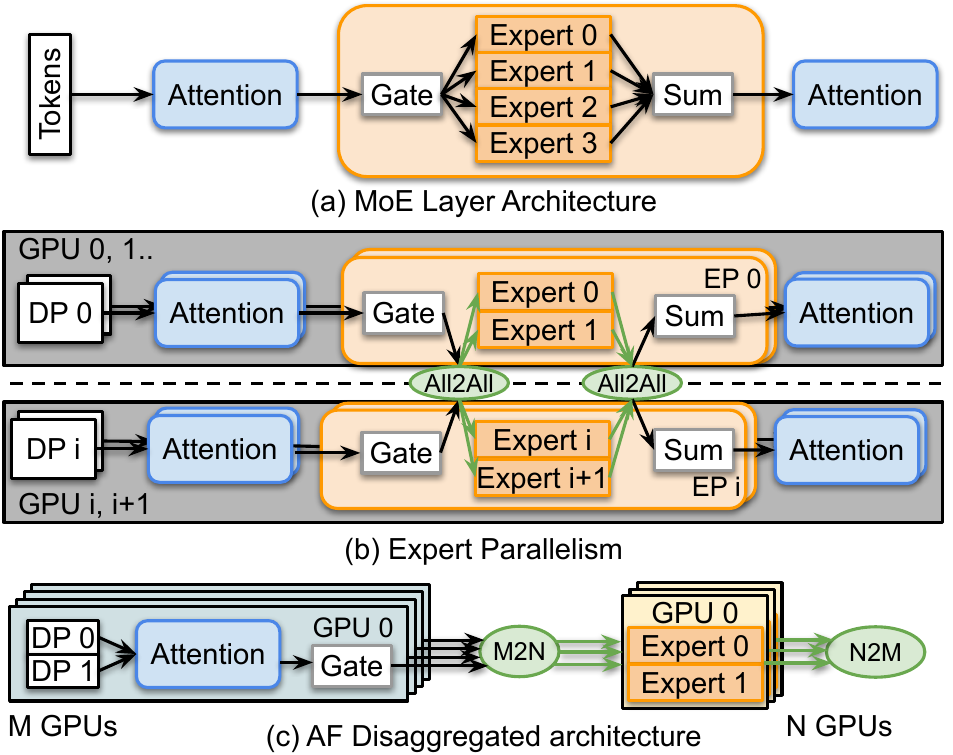}
    \caption{Overview of MoE architecture, EP, and Attention-FFN Disaggregated architecture.}
    \label{fig:moe_overview}
\end{figure}

As shown in Figure \ref{fig:moe_overview} (a), MoE models use a trainable gate network to select the \textit{top-k} experts with the highest probability scores.
The outputs of these active experts are then reduced by a weighted sum, and the result is passed as the input to the next attention layer. Modern large-scale MoE models exhibit very sparse activation, whose principle of conditional computation facilitates the scaling of model parameters to hundreds of billions while incurring only a sub-linear increase in floating-point operations (FLOPs). For example, the DeepSeek-V4-Pro model \cite{deepseek_v4_pro}, with 1.6T total parameters, activates 49B for each token by engaging 6 out of 384 total experts.

\subsection{Distributed LLM Training}

Distributed training scales LLMs beyond single-device limits through complementary parallelization strategies.

\textbf{Data Parallelism (DP).} DP replicates model states across devices and splits batches for independent computation, synchronizing gradients via all-reduce. Memory-efficient variants like DeepSpeed-ZeRO \cite{rajbhandari2020zero} and FSDP \cite{zhao2023pytorch} shard parameters, gradients, and optimizer states.

\textbf{Pipeline Parallelism (PP).} PP divides layers into sequential stages connected by activation transfers. Megatron-LM \cite{shoeybi2019megatron, narayanan2021efficient} adopts a 1F1B schedule to reduce activation memory, leaving only minor pipeline bubbles.

\textbf{Expert Parallelism (EP).} MoE uses EP to shard experts across GPUs, with tokens dynamically routed to their assigned experts (Figure~\ref{fig:moe_overview}(b)). Each forward/backward pass includes two all-to-all operations, \textit{dispatch} and \textit{combine}, for expert communication.

\begin{table}[t]
\centering
\caption{Description of \name symbols.}
\label{tab:symbols}
\resizebox{\columnwidth}{!}{%
\begin{tabular}{ll}
\hline
\textbf{Symbol} & \textbf{Description} \\
\hline
$L$   & Number of transformer layers \\
$E$   & Total number of experts \\
$k$   & Top-K experts that each token routed\\
$H$   & Hidden size \\
$D_e$ & Hidden size of the FFN layer in experts \\
$s$   & Input sequence length \\
$b$   & Batch size \\
$W$   & Total parallel world size\\
$m,M$ & Disagg. Attention nodes/GPUs number \\
$n,N$ & Disagg. FFN nodes/GPUs number \\
$EP$  & Expert parallel size \\
$P$   & GPU compute density (FLOPs/s) \\
$B_{\text{NV}}, \,B_{\text{IB}}$ & NVLink / IB bandwidth (Bytes/s)\\
$p$   & Groups per component (pipeline depth) \\
$v$ & Virtual stages per group \\
\hline
\end{tabular}%
}
\end{table}

\section{Challenges and Motivations}

In this section, we first illustrate the dominant communication overhead under large-scale MoE training and explains the inadequate overlapping from existing methods. Then we introduce how these system-level problems motivate AF disaggregation.

\subsection{Dominant MoE Communications Overhead}
\label{sec:largea2a}

In large-scale EP training \cite{deepseekai2025deepseekv3technicalreport, yang2025qwen3}, experts are sharded across multiple GPUs and nodes. Using the notation in Table \ref{tab:modelconfig}, the expected per-GPU all-to-all communication volume is \(V=\frac{EP-1}{EP}Nk\), since each GPU exchanges tokens with every non-local GPU. Transfers use NVLink within a node and InfiniBand (IB) or Ethernet across nodes \cite{gangidi2024rdma}. End-to-end comunication latency is dominated by the slower interconnect, typically IB near 100GB/s, well below NVLink near 900GB/s. As the EP size spans more nodes, a larger share of peers is reached over IB, which increases communication latency overhead.

\begin{figure}[h]
    \centering
    \begin{subfigure}[b]{0.48\linewidth}
        \centering
        \includegraphics[width=\linewidth]{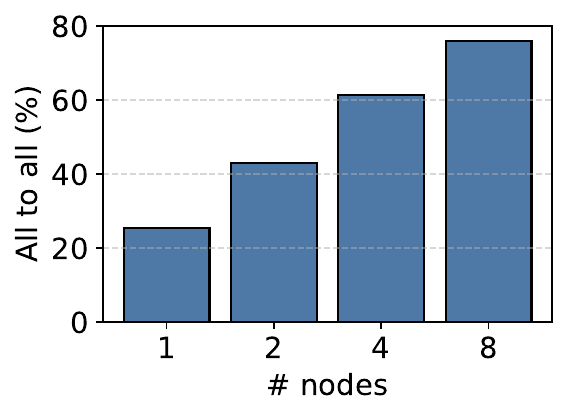}
        \caption{All-to-all percentage over total latency, \(topk=4\)}
        \label{fig:all2all:a}
    \end{subfigure}
    \hfill
    \begin{subfigure}[b]{0.48\linewidth}
        \centering
        \includegraphics[width=\linewidth]{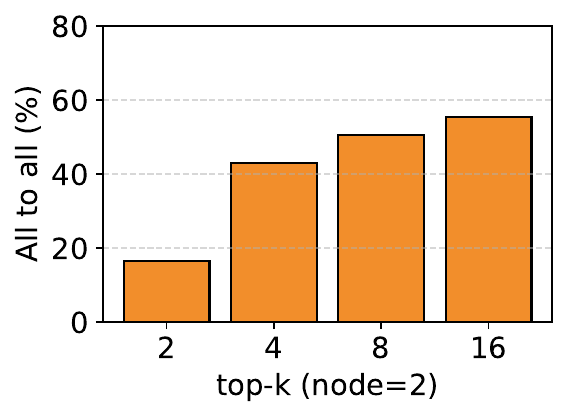}
        \caption{All-to-all percentage, two-node H800}
        \label{fig:all2all:b}
    \end{subfigure}
    \caption{(a) MoE all-to-all share (\%) per layer across nodes. (b) effect of varying top-\(k\) selection. Both under 8K sequence length in DeepSeek-MoE.}
    \label{fig:all2all}
\end{figure}

Figure \ref{fig:all2all:a} shows that as the EP degree spans more nodes, the all-to-all share of total training time rises from 22\% on a single node to nearly 78\% on 8~nodes. In parallel, since the communication volume \(V\) grows almost linearly with the number of top-k selected tokens, as shown in Figure \ref{fig:all2all:b}, the percentage of all-to-all increases a lot. Scaling along either axis therefore incurs substantial communication overhead during training. As communication comes to dominate, compute resources idle until exchanges finish, pushing the system into a communication-bound regime.

\subsection{Unlocked operation-level overlapping space}
\label{sec:op}
Given the dominant all-to-all overhead and the two symmetric all-to-all phases (dispatch and combine) around attention and FFN, prior systems \cite{tutel, zhang2025comet, jiang2024lancet, he2022fastermoe} attempt to hide communication by slicing the computation into operation-level chunks and pipelining dispatch and combine with computation across chunks, thereby overlapping communication with compute.

\begin{figure}[t]
    \centering
    \includegraphics[width=\linewidth]{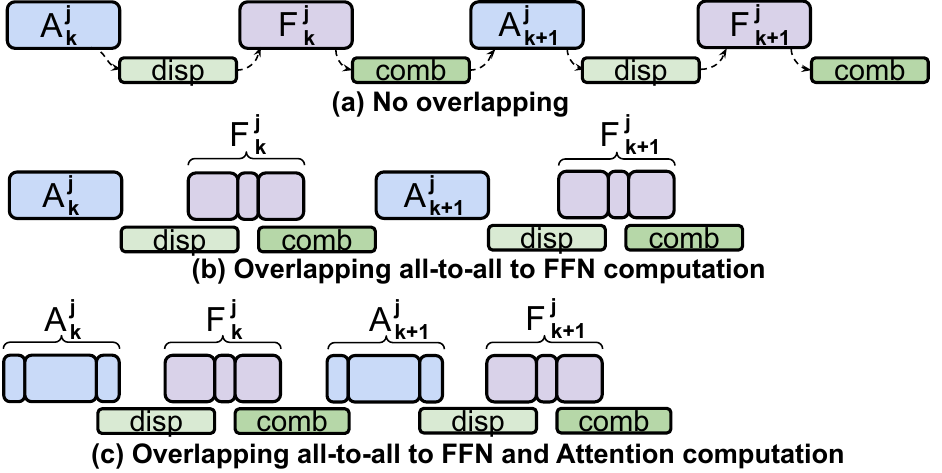}
\caption{(a) Vanilla MoE forward. (b) Overlap all-to-all with FFN, Comet and Tutel. (c) microbatch-level pipelining to reduce bubbles. Non-overlapped tails remain when all-to-all dominates.}
    \label{fig:motivationchunk}
\end{figure}

\begin{figure}[h]
    \centering
\includegraphics[width=\linewidth]{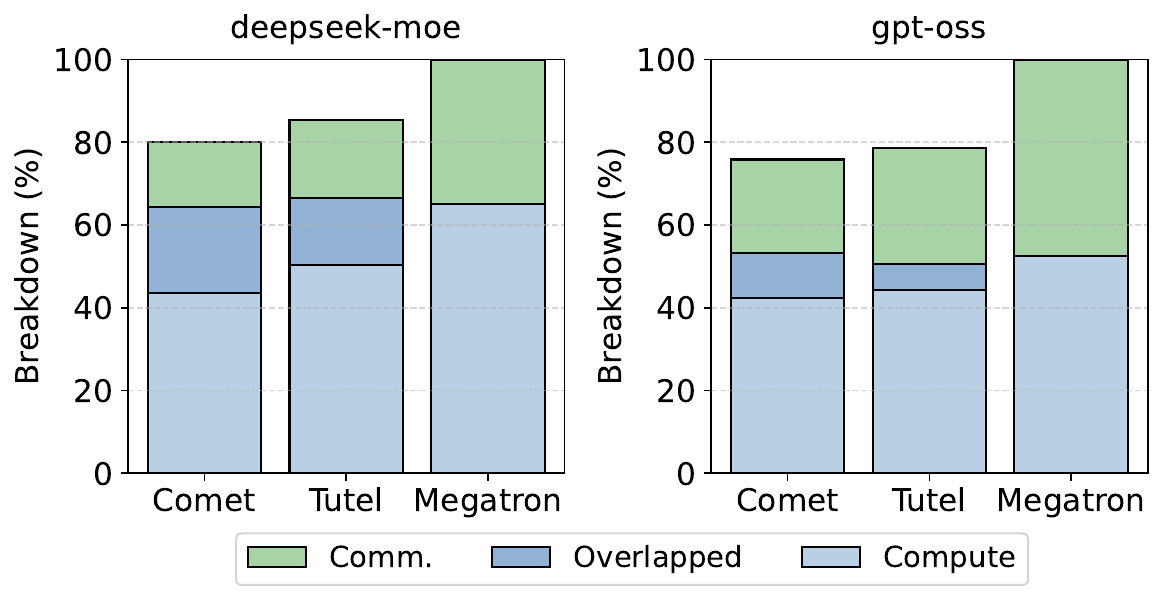}
    \caption{Under operation-level overlap methods, it remains communication tails that cannot be hidden with compute.}
    \label{fig:motivation_comet}
\end{figure}

Since GroupGEMM in the FFN admits natural chunking, tokens can be transmitted and received in chunks and the compute can overlap with chunked communication. Hence, most prior systems \cite{tutel, zhang2025comet, he2022fastermoe} confine overlap to the all-to-all phase and the FFN compute, as shown in Figure \ref{fig:motivationchunk}. A compiler-driven approach \cite{jiang2024lancet} extends overlap by microbatch-level partitioning and pipelining to shrink the communication bubbles around attention. Yet self-attention with FlashAttention \cite{dao2022flashattention} remains hard to chunk completely, so when all-to-all dominates there are still non-overlapped communication tails, as illustrated in Figure \ref{fig:motivation_comet}.

We profiled the operation-level overlap and found that the narrow focus adopted by prior work constrains overlap opportunities and yields suboptimal performance in Figure \ref{fig:motivation_comet}. These methods emphasize per-operator overlap while overlooking the richer module-level overlap that spans attention, and FFN. This insight motivates a more general scheduling strategy to unlock the full module-level overlapping space.

\subsection{Compute-Communication imbalance between attention and FFN}
\label{sec:mot_roof}
Consider hybrid EP–DP parallelism across nodes, the most common configuration used in real-world~\cite{ yang2025qwen3}. For each EP group, the attention FLOPs cost is $\mathrm{F}_{\text{a}}(s)=\alpha_1 s^2+\alpha_2 s$, and the FFN is $\mathrm{F}_{\text{f}}(s)=\beta s$, where \(\alpha_1,\alpha_2,\beta\) are constants defined by model dimensions such as hidden size, detailed in Sec. \ref{sec:allocation}. On the networking side, both dispatch and combine all-to-all phases transfer the same token volume, giving a total $V=\gamma s$, where \(\gamma\) depends on model parameters such as hidden width and top-\(k\).

\begin{figure}[t]
    \centering
    \begin{subfigure}[h]{0.49\linewidth}
        \centering
        \includegraphics[width=0.9\linewidth]{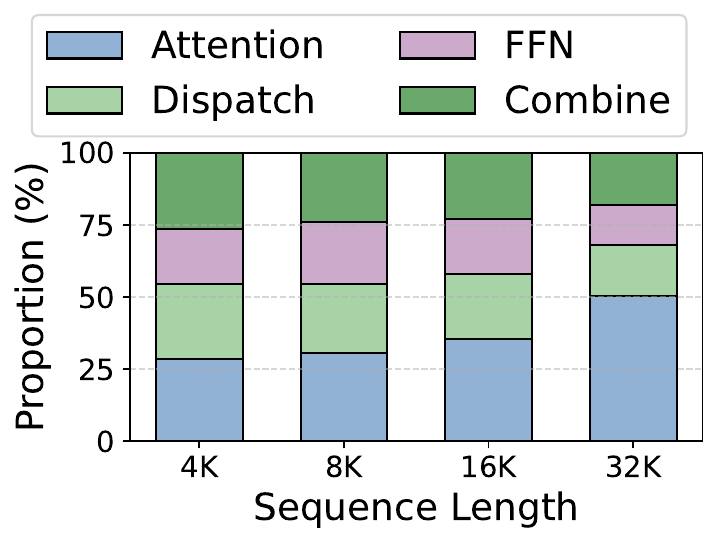}
        \caption{Latency percentage of components.}
        \label{fig:roofmotivation:a}
    \end{subfigure}
    \hfill
    \begin{subfigure}[h]{0.49\linewidth}
        \centering
        \includegraphics[width=\linewidth]{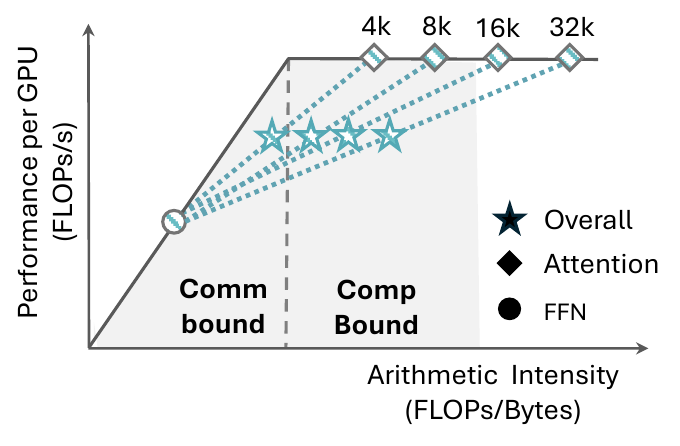}
        \caption{Compute-Communication roofline model.}
        \label{fig:roofmotivation:b}
    \end{subfigure}
    \caption{(a) Latency breakdown of attention, FFN, and all-to-all communication when trainining Qwen-3 across 8 nodes. (b) Compute-Communication roofline showing attention and FFN under the same network roof. The overall point, as a weighted average of both modules, illustrates their imbalance and motivates disaggregated resource allocation.
    }
    \label{fig:roofmotivation}
\end{figure}

As shown in Figure~\ref{fig:roofmotivation:a}, within the same EP group, the attention component’s share of total computation increases from 28.38\% to 50.26\% as the sequence length grows from 4K to 32K, while the FFN share decreases from 19.22\% to 13.96\%. The compute-to-communication ratio for attention rises from 1.08 to 2.78, whereas that for FFN remains nearly constant, changing slightly from 0.73 to 0.77. These trends highlight the distinct network–computation characteristics of attention and FFN layers.

We also apply the roofline model \cite{scaling-book} to characterize the compute–communication behavior of attention and FFN, as shown in Figure~\ref{fig:roofmotivation:b}. A system is \textit{network-bound} when data transfer is slower than computation (\(T_{\text{comm}} > T_{\text{comp}}\)), making bandwidth the bottleneck, and \textit{compute-bound} when performance is limited by peak FLOPs throughput. The tuning point is defined as \(\hat{I} = \frac{\text{Peak FLOPs}}{\text{Peak Bandwidth (Bytes/s)}}\), with arithmetic intensity \(I = \frac{\text{Compute FLOPs}}{\text{Communication Bytes}}\).

Since attention computation grows quadratically with sequence length while FFN and communication scale linearly, attention’s arithmetic intensity increases faster, allowing it to reach the compute-bound regime earlier. Within each EP group, dispatch and combine operations transfer equal data volumes, so \(I_a\) (attention) surpasses \(\hat{I}\) before \(I_f\) (FFN) as sequence length increases.

This divergence highlights a fundamental imbalance: attention efficiently utilizes compute, whereas FFN remains constrained by communication. As a result, overall system performance (can be viewed as a weighted one between attention and FFN) is bounded by FFN’s communication overhead. These insights motivate a disaggregated architecture that separates attention and FFN into independent resource groups, allowing each to operate near its optimal compute–communication balance.
\section{Design of \name}
In this section, we present \textsc{\name}, a training system for large-scale MoE models that mitigates all-to-all bottlenecks by jointly rethinking placement and schedule in a disaggregated design. We organize the design in dependency order: a \textit{disaggregated placement} (Section~\ref{sec:Disaggregatedcomponent}) that maps attention and FFN to disjoint workers, a throughput-oriented pipeline schedule \textit{AF-Pipe} (Section~\ref{sec:afpipe}) built on this placement for training, and an \textit{adaptive worker allocation} (Section~\ref{sec:allocation}) that efficiently searches over pipeline configurations.

\subsection{Disaggregated component placement}
\label{sec:Disaggregatedcomponent}

We first introduce the disaggregated architecture underlying \name. In the disaggregated architecture, attention and FFN layers are partitioned into distinct groups and assigned to separate worker sets. Specifically, \name partitions the Transformer by \textit{component type} and, for each type, forms interleaved
\textit{groups}. We use a unified notation
\(c\in\{A,F\}\), where \(A\) denotes self-attention and \(F\) denotes the feed-forward
block.

Let the model have \(L\) layers indexed \(0,\ldots,L-1\). Fix \(p\ge 1\) groups per
component type. For group \(g\in\{0,\ldots,p-1\}\) of component \(c\), the assigned
layers are
\begin{equation}
\mathcal{L}^{c}_{g}
=\{\, g + k p \;\mid\; \ k = 0, 1, \dots, v-1 \,\}.
\end{equation}
Thus each group receives every \(p\)-th occurrence of its component across depth.
(When \(L\) is divisible by \(p\), we write \(v=L/p\) and the enumeration truncates
at \(k=v-1\); otherwise, group sizes differ by at most one.) Unless noted, we use the
same \(p\) for \(A\) and \(F\).

\begin{figure}[h]
    \centering
    \includegraphics[width=1.0\linewidth]{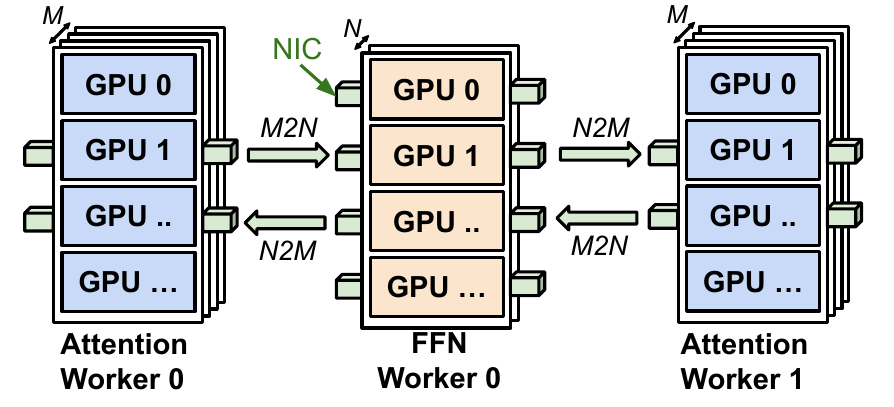}
    \caption{Disaggregated attention–FFN placement. Attention groups (\(\{A_0,A_2,A_4,A_6\}\), etc.) are assigned to A-Workers, where GPUs are replicated using data parallelism (DP). FFN groups are placed on F-Workers, where experts are distributed across GPUs following expert parallelism (EP). Workers communicate through many-to-many (M2N/N2M) links via NICs, enabling overlapped compute and communication across modules.}
    \label{fig:m2n}
\end{figure}

\textit{Example.} With \(L{=}8\) and \(p{=}2\), attention groups are
\(\{A_0,A_2,A_4,A_6\}\) and \(\{A_1,A_3,A_5,A_7\}\); the \(F\) groups follow the
same pattern. Each group \(g\) of type \(c\) is placed on a dedicated worker
\(W^{c}_{g}\). The output embedding of the final layer resides on the worker hosting
that layer’s \(F\) block.

For the attention workers (A-Workers), GPUs within each group are replicated following data parallelism (DP). For the FFN workers (F-Workers), experts within each FFN layer are distributed across GPUs inside the group using expert parallelism (EP). This design keeps all groups homogeneous and identically configured within each component type, enabling larger MoE configurations under fixed device-memory budgets and allowing uniform scaling across groups, thereby maintaining balanced compute and communication utilization as analyzed in Section~\ref{sec:allocation}.

\begin{figure*}[!t]
    \centering
    \includegraphics[width=1.0\linewidth]{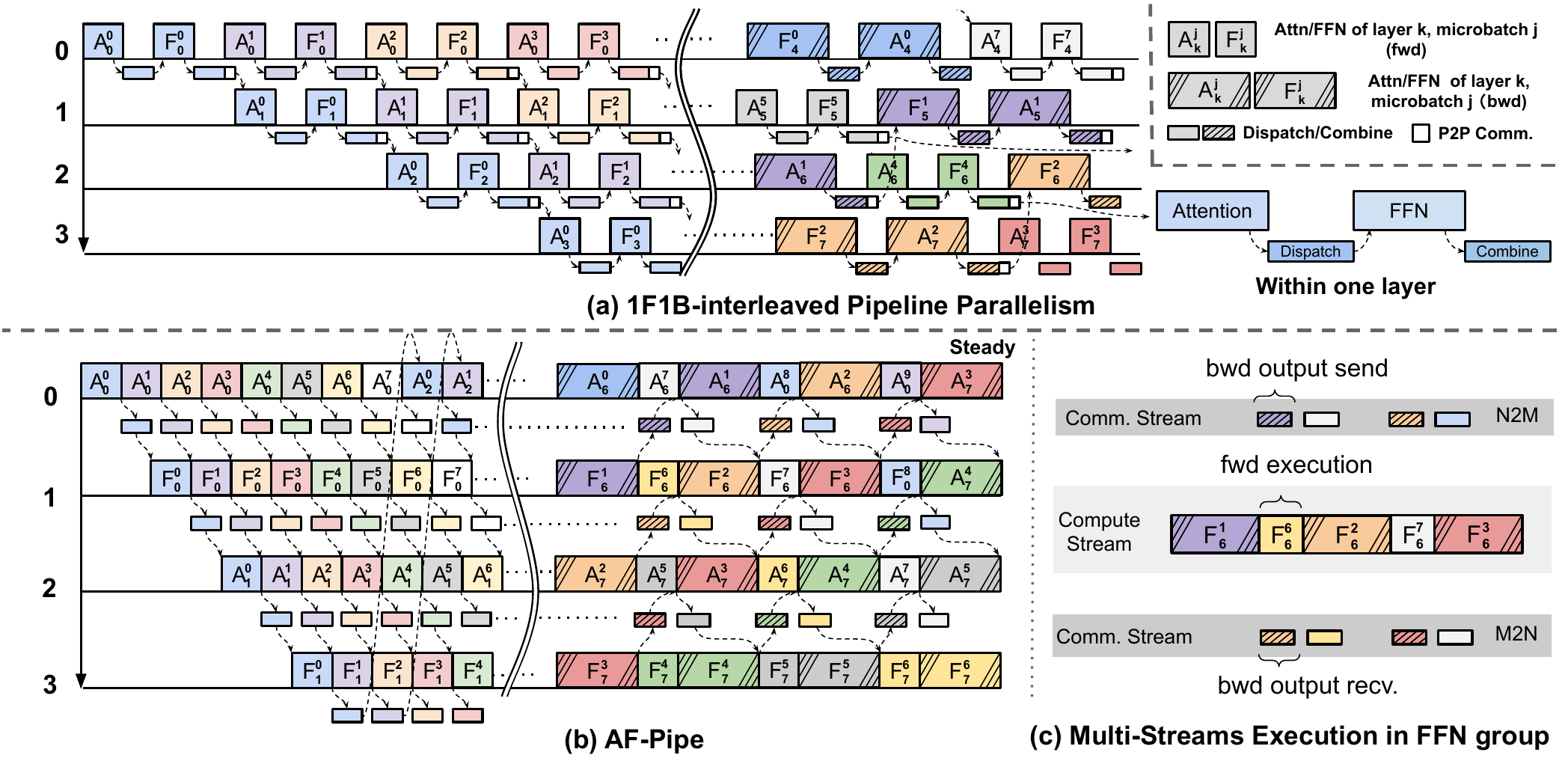}
    \caption{(a) Naive 1F1B pipeline: overlaps inter-stage communication but leaves all-to-all unhidden. (b) AF-Pipe overview: treats MoE communication as a first-class pipeline stage. (c) Multi-stream execution: overlaps computation and communication across streams.}
    \label{fig:afpipe}
\end{figure*}
\subsection{AF-Pipe with overlapping}
\label{sec:afpipe}

In this section, we describe how to efficiently train disaggregated MoE-based LLMs at scale using a unified multi-stage pipeline with overlapping, termed \textbf{AF-Pipe}.

\subsubsection{AF-Pipe Time-multiplexing scheduling}

Since attention and FFN modules are partitioned into separate worker groups, training a single batch independently would cause idle periods for one group while the other is active. To mitigate this inefficiency, as illustrated in Figure~\ref{fig:afpipe}, \mbox{\name} introduces AF-Pipe, a tailored pipeline parallelism (PP) strategy that unifies disaggregated architecture with time-multiplexing-level pipeline parallelism for efficient large-scale MoE training.

Specifically, AF-Pipe adopts the AF disaggregated architecture described in Section~\ref{sec:Disaggregatedcomponent}, where each worker executes only one component type, either attention or FFN. During training, a batch is divided into multiple microbatches, and execution is pipelined across them. Workers perform both forward and backward passes, exhibiting bidirectional behavior. In the forward pass, intermediate results (the hidden states) of each microbatch are transferred between adjacent stages, from the attention to the FFN workers. In the backward pass, the gradients of \textit{hidden\_states} propagate in the reverse direction, from FFN back to attention workers. This design ensures both component groups remain continuously active, achieving efficient utilization and minimizing pipeline idle time.

\subsubsection{AF-Pipe with M2N stage boundaries across groups}

In \name, since each worker hosts the same component and assigned with different GPU resources for different types (assuming M for A-Worker and N for F-Worker). The traditional all-to-all exchanges evolve into \textit{many-to-many} communication primitives. AF-Pipe introduces multi-stage \textit{many-to-many} (M2N/N2M) communication boundaries between attention and FFN groups to eliminate redundant data transfers and reduce large-expert communication overhead, extending ideas from MegaScale-Infer~\cite{zhu2025megascale} and StepMesh~\cite{wang2025step}. This multi-stages M2N communication will behave full-duplex so the bandwidth can be fully utilized. By fusing conventional point-to-point (P2P) and combine operations into a single unified communication stage, AF-Pipe reduces overall communication cost by roughly \(1/k\) and enables systematic computation–communication overlap across groups. This design keeps both attention and FFN workers highly utilized while effectively hiding large-scale token dispatch and combine latency behind computation. As illustrated in Figure~\ref{fig:afpipe}(b), each transformer layer includes two major all-to-all communications between attention and FFN. In conventional setups, tokens from FFN must first perform a combine step before initiating inter-stage P2P exchanges, creating redundant data transfers and additional latency.

We compare the warmup bubbles of AF-Pipe and standard pipeline parallelism. For the baseline:
\begin{equation}
B_{\text{base}} = \frac{(PP-1)\,[L(T_a + T_f + 2T_{a2a}) + T_{\text{p2p}}]}{v},
\end{equation}
where \(T_a\), \(T_f\), and \(T_{a2a}\) are attention, FFN, and all-to-all times per layer.
In AF-Pipe, inter-stage communication is larger but fully overlapped via M2N exchange:
\begin{equation}
B_{\text{AF}} = \frac{(PP-1)\,[\max(T_a,\,T_f)+T_{\text{M2N}}]}{2v}.
\end{equation}
With per-layer staging (\(L=1\)) and \(T_{\text{M2N}}\!\approx\!T_{a2a}\), AF-Pipe’s bubble is about one-fourth of the baseline, as it removes P2P delay and fuses two all-to-all operations into one overlapped M2N stage.

\subsubsection{M2N communication overlapped with asynchronous execution}

In the AF-Pipe, the traditional all-to-all exchanges are replaced with \textit{many-to-many} (M2N/N2M) communication between attention and FFN groups to eliminate redundant transfers. We then describe how M2N overlaps with computation by asynchronous execution streams to maximize utilization.

As illustrated in Figure~\ref{fig:afpipe}, each A-group and F-group maintains dedicated send and receive \textit{ProcessGroups}, launched asynchronously across three coordinated streams in each workers, forward, backward, and communication, to guarantee non-blocking execution. For example, the steady-state schedule interleaves computation and communication across streams. At time period of \(T_0\), the F-worker \textit{sends (N2M)} the gradients results of backward computation \(F^1_6\) to the previous A-worker , while the compute stream concurrently executes the forward pass of \(F^6_6\) on already available data.  At the same time, this F-worker \textit{receives} the backward results from the next A-worker \(A_7^2\). Through careful resource allocation (Section \ref{sec:allocation}), the compute duration of each stage remains balanced, maintaining high pipeline utilization and minimizing idle time. Symmetrically, the A-worker behave the same asynchronous execution. This design not only overlaps computation and communication but also minimizes synchronization overhead, achieving near-continuous GPU utilization across the entire pipeline.

\subsection{Adaptive worker allocation}
\label{sec:allocation}
This section presents a Compute-Communication Roofline model to guide the adaptive allocation of disaggregated compute and communication resources, and casts the GPU/NIC partitioning between attention and FFN groups as a Mixed Integer Linear Programming (MILP) problem. Unlike serving, where input shapes are dynamic and latency-bound, pretraining workloads run with fixed sequence lengths and batch sizes that are known offline~\cite{qwen35blog, yang2025qwen3, deepseek_v4_pro, zai_glm51}, so a one-shot static allocation suffices to balance throughput and utilization across stages.

\begin{figure}[t]
    \centering
    \includegraphics[width=0.85\linewidth]{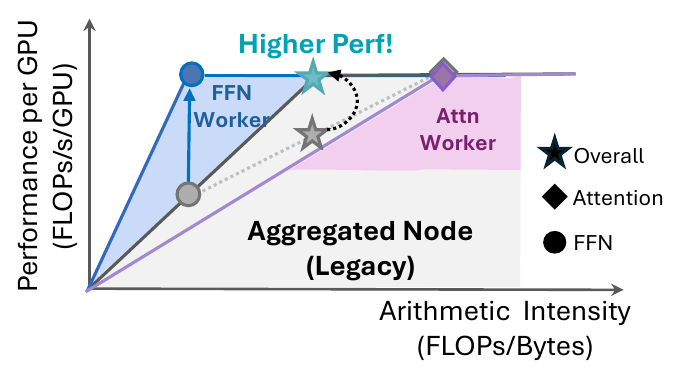}
    \caption{Compute-Communication roofline comparison of aggregated and disaggregated (AFD) architectures. In the aggregated baseline (gray), attention and FFN share the same slope (network bandwidth) and roof (compute), leaving FFN communication-bound and depressing the overall operating point. In AFD, we can redistributes NIC bandwidth across worker groups via resource allocation. FFN workers (blue) receive higher effective per-GPU network bandwidth (steeper slope) and move closer to the compute roof, while attention workers (purple) remain near the compute roof, yielding a higher overall system performance.
}

    \label{fig:dis_roofline}
\end{figure}

\noindent\textbf{Compute-Communication Roofline Model for AFD.}
\label{sec:ana}
To capture the imbalance compute–communication characteristics of attention and FFN described in Section~\ref{sec:mot_roof}, we extend the classical roofline model to the AFD architecture. Notably, in large EP, we only consider inter-node communication, e.g., IB bandwidth. In AF-Pipe (Section~\ref{sec:afpipe}), attention and FFN execute in a time-multiplexed pipeline and communicate bidirectionally between groups. For simplicity, we model the system as two communicating worker clusters: attention (\(m\) nodes) and FFN (\(n\) nodes). We fix the number of NICs for A/F- worker and ensure both groups collectively occupy half of the system’s total NICs, maintain balanced ingress/egress bandwidth, avoiding link imbalance during inter-group communication.

Following Section~\ref{sec:mot_roof}, the aggregated system turning point is determined by the arithmetic intensity threshold \(\hat{I}\). Each group’s peak FLOPs scale with its GPU count (\(m\) for attention, \(n\) for FFN), while both sustain identical peak network bandwidth through balanced NIC allocation. The effective turning points differ as:
\begin{equation}
    \hat{I}^{\text{A}} = \frac{2m}{m+n} \hat{I}, \qquad
    \hat{I}^{\text{F}} = \frac{2n}{m+n} \hat{I}.
\end{equation}

We focus on dominant GEMM operations and ignore minor non-linear terms. Taking group-query attention with group size \(g\) as an example (parameters in Table~\ref{tab:symbols}), the attention part includes hidden-state projection to QKV and self-attention, whose FLOPs are approximately \(C_{\mathrm{a}}=b(SH^{2}(2+2/g)+4S^{2}H)\). For the expert part, after gating \(kS\) tokens, two GroupGEMMs (up- and down-projection) require about \(C_{\mathrm{f}}=b(4kSHD_e)\) operations. The total communication volume for bfloat16 data is roughly \(b(2SkH)\). Consequently, the arithmetic intensities are \(I_{\text{attn}}=\frac{H(2+2/g)+4S}{2k}\) and \(I_{\text{ffn}}=2D_e\), while the system’s turning point \(I_m\). Figure~\ref{fig:dis_roofline} illustrates how the per-GPU performance scales with arithmetic intensity under the compute–communication roofline model.

In legacy nodes, attention and FFN share identical per-GPU network bandwidth (the slope) and peak GPU FLOPs (the compute roof). The slope corresponds to the per-GPU InfiniBand bandwidth, and the overall system performance lies along the line segment connecting the two components’ operating points—below the compute roof. In the AFD architecture, since \(\hat{I}^{\text{A}}\) and \(\hat{I}^{\text{F}}\) differ, we can lift the FFN group’s communication roof (slope) and drive its compute closer to the peak roof. Meanwhile, \(\hat{I}^{\text{F}}\) shifts rightward, while the attention group still reaches the compute roof. As shown in Figure~\ref{fig:dis_roofline}, FFN nodes require higher InfiniBand bandwidth to fully utilize their compute capacity, whereas attention nodes are more compute-bound and thus demand less bandwidth per GPU.

\noindent\textbf{Roofline-model guided resource placement.}
We jointly split $W$ GPUs and $M_{\mathrm{tot}}$ NICs between the two groups: $M=m\mu$ GPUs to attention and $N=n\nu$ to FFN with NIC budget. Each stage's latency is the max of its compute and communication time,
\begin{align}
\mathcal{T}_a &= \max\!\Big(\tfrac{C_a}{PM},\ \tfrac{V}{M_a B_{\mathrm{IB}}}\Big), \\
\mathcal{T}_f &= \max\!\Big(\tfrac{C_f}{PN},\ \tfrac{V}{M_f B_{\mathrm{IB}}}\Big),
\end{align}
where $C_a, C_f$ are per-iteration FLOPs and $V$ is the inter-group volume.

AF-Pipe couples the two groups as a producer--consumer pipeline, so iteration time tracks $\max(\mathcal{T}_a,\mathcal{T}_f)$ and any imbalance becomes a bubble; maximizing aggregate intensity $I_{\mathrm{attn}}+I_{\mathrm{ffn}}$ alone is therefore \textit{not} sufficient. We decompose the placement: (i) the GPU split sets $T^\star=\min\max(\mathcal{T}_a,\mathcal{T}_f)$, removing the bubble-generating bottleneck; (ii) among configurations attaining $T^\star$, the NIC split maximizes $\mathrm{MFU}=(C_a{+}C_f)/(T^\star W P)$, equivalent under fixed $T^\star$ to maximizing $I_{\mathrm{attn}}+I_{\mathrm{ffn}}$, by lifting the compute roof on the communication-slack side:

\begin{subequations}
\label{eq:opt}
\vspace{-5mm}
\small
\begin{align}
T^\star=\min\ & \max(\mathcal{T}_a,\mathcal{T}_f), \label{eq:opt-a}\\
\max\ & I_{\mathrm{attn}}+I_{\mathrm{ffn}}\quad \text{s.t.}\ \max(\mathcal{T}_a,\mathcal{T}_f)\le T^\star, \label{eq:opt-b}\\
\text{s.t.}\ & M\!+\!N\!=\!W,\ M_a\!+\!M_f\!=\!M_{\mathrm{tot}}, \nonumber \\
& m,n,M_a,M_f\!\ge\!1,\ \mu,\nu\!\in\![8]. \nonumber
\end{align}
\end{subequations}
The MILP based on \cite{gurobi} yields a roofline-guided seed $(M_0, M_{a,0})$, which Algorithm~\ref{alg1} refines via local search on measured per-step times to absorb effects unmodeled by the analytic costs.

\begin{algorithm}[h]
\small
\caption{AFD allocation: roofline seed with profile-guided refinement.}
\label{alg1}
\begin{algorithmic}[1]
\Require $\Theta=(C_a,C_f,V,P,B_{\mathrm{IB}})$, $W$, $M_{\mathrm{tot}}$, search radius $r$, trials $K$, tolerance $\epsilon$
\Ensure GPU split $M^\star$, NIC split $M_a^\star$
\State $\mathcal{S}\!\gets\!\emptyset$, $\mathcal{T}^\star\!\gets\!\infty$ \Comment{Phase~1: minimize bottleneck stage via Eq.~\eqref{eq:opt-a}}
\For{feasible $(M,M_a)$ under Eq.~\eqref{eq:opt}}
  \State $\mathcal{T}\!\gets\!\max(\mathcal{T}_a,\mathcal{T}_f)$
  \If{$\mathcal{T}<\mathcal{T}^\star\!-\!\epsilon$} $\mathcal{T}^\star\!\gets\!\mathcal{T}$, $\mathcal{S}\!\gets\!\{(M,M_a)\}$
  \ElsIf{$\mathcal{T}\le\mathcal{T}^\star\!+\!\epsilon$} $\mathcal{S}\!\gets\!\mathcal{S}\cup\{(M,M_a)\}$ \EndIf
\EndFor
\State $(M_0,M_{a,0})\!\gets\!\arg\max_{\mathcal{S}}(I_{\mathrm{attn}}\!+\!I_{\mathrm{ffn}})$ \Comment{Phase~2: MFU tie-break (Eq.~\eqref{eq:opt-b})}
\State $(M^\star,M_a^\star,T^\star)\!\gets\!(M_0,M_{a,0},\mathrm{Profile}(M_0,M_{a,0}))$ \Comment{Phase~3: local refinement}
\For{$k\!=\!1$ to $K$}
  \State $(M',M_a')\!\gets\!\mathrm{clip}((M^\star,M_a^\star)+\mathrm{rand}(-r,r))$
  \If{$\mathrm{Profile}(M',M_a')<T^\star$} update $(M^\star,M_a^\star,T^\star)$ \EndIf
\EndFor
\State \Return $(M^\star,M_a^\star)$
\end{algorithmic}
\end{algorithm}

\section{Implementation}
We implement \name in 6K lines of Python and 2K C++, with components with Megatron-LM based on PyTorch v2.6 for NVIDIA GPUs. For efficient dispatching and combine data between AF, we implement an M-to-N/N-to-M communication primitives supported by GPUDirect \cite{nvidia_gpudirect_2025} and GPUCopy \cite{nvidia_gdrcopy_2025}, following the idea of \cite{zhu2025megascale, wang2025step}. In \name, we only change the module-level definition logits, trainers only need to define the attention and FFN computation as usual, and AF-Pipe will partition the transformer blocks into groups. \name is seamlessly integrated into Megatron-LM \cite{shoeybi2019megatron} to reuse the other model architecture, 3D parallelism, and hyperparameters.
\section{Evaluation}
\subsection{Setups}
\noindent\textbf{Testbed.} We deploy \name on the cluster, which consists of 16 nodes, equipped with 8 Nvidia H800 GPUs (80 GB memory each), 168 CPUs and equipped with ConnectX-7 8×400 GbE NICs for each node. GPUs within the same node are interconnected via 400GB/s NVLINK.

\begin{table}[h]
\centering
\caption{MoE model configurations}
\label{tab:modelconfig}
\resizebox{\columnwidth}{!}{%
\begin{tabular}{lccccc}
\toprule
\textbf{Model} & \textbf{Layers} & \textbf{Hidden} & \textbf{Experts} & \textbf{top-k} & \textbf{MoE Hidd.} \\
\midrule
Deepseek-moe & 28 & 2048 & 64  & 4 & 1408 \\
GPT-OSS      & 36 & 2880 & 128 & 4 & 2880 \\
Qwen3        & 94 & 4096 & 128 & 8 & 1536 \\
\bottomrule
\end{tabular}}
\end{table}

\noindent\textbf{Models.} We evaluate \name based on three MoE models, DeepSeek-MoE \cite{dai2024deepseekmoe}, GPT-OSS-120B \cite{agarwal2025gpt}, and Qwen3-235B-A22B \cite{yang2025qwen3}. The model parameter details are listed in Table \ref{tab:modelconfig}. We explore the number of experts on different depths (layers) and widths (hidden states size) in our system. The global batch size is set to be larger than on the pipeline stages to amortize the pipeline bubbles.

\begin{figure*}[t!]
    \centering
    \includegraphics[width=0.8\linewidth]{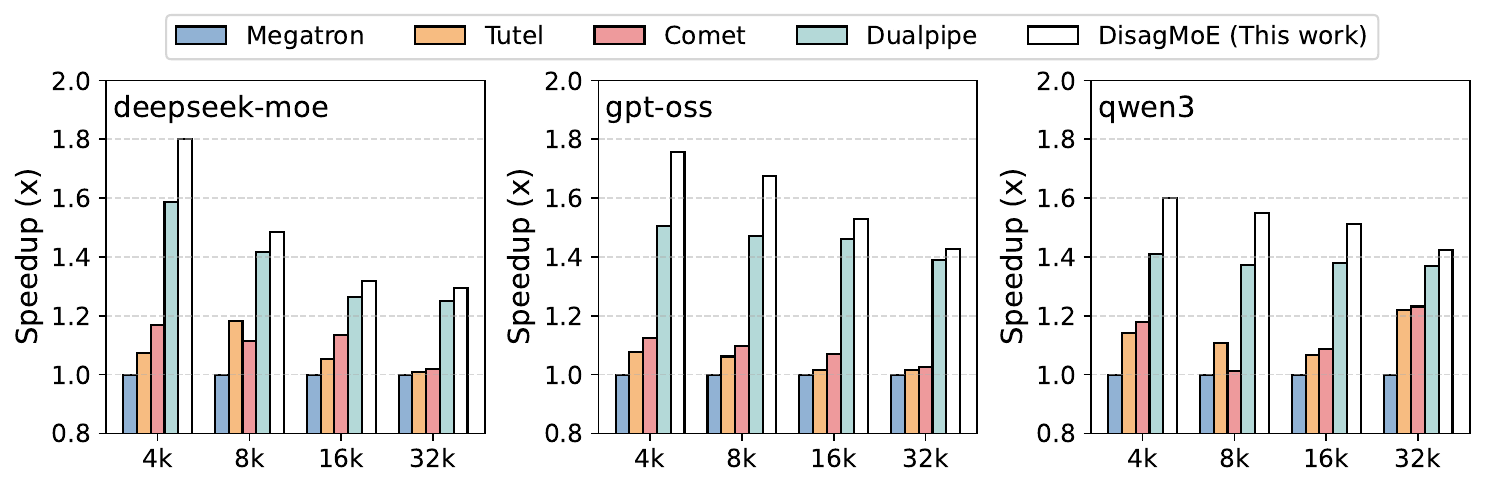}
    \caption{End-to-end training throughput across different sequence lengths (4K–32K) on DeepSeek-MoE, GPT-OSS, and Qwen3. \name consistently outperforms Megatron, Tutel, Comet, and DualPipe, achieving up to 1.81× speedup over Megatron-1F1B and 1.34× over state-of-the-art MoE overlap systems.}
    \label{fig:overall}
\end{figure*}

\subsection{End-to-end Performance}

\noindent\textbf{Baseline.} We compare \name with four SOTA training methods: (1) Megatron-LM: a widely-used open-source training framework from NVIDIA with 3D parallelism (2) Tutel: the state-of-the-art MoE overlapping optimizations \cite{tutel}, (3) Comet \cite{zhang2025comet}: a fine-grained communication overlapping library for MoE (4) DualPipe \cite{dai2024deepseekmoe}: a bidirectional pipeline parallelism overlap of forward and backward computation-communication.

In this subsection, we evaluate the training throughput of our testbed across sequence lengths ranging from 4K to 32K. We compare \name’s throughput with multiple baselines under varying numbers of GPUs. To ensure fairness, we fix the local batch size (\textit{i.e.,} per-device batch size), allowing the model’s effective total batch size to scale linearly with the number of GPUs. For all baselines, we set the expert parallelism (EP) size to 16 across two nodes and carefully tune the pipeline and virtual stage configurations to avoid OOM errors. For \name, we set the optimal resource allocation based on Sec. \ref{sec:allocation} .

The overall results are presented in Figure \ref{fig:overall}. As shown, \name consistently outperforms all baselines, achieving up to 1.81× speedup over the naïve Megatron-1F1B interleaved overlap and 1.34× speedup compared to the SOTA overlapping training methods. We provide a detailed analysis of these results in the following paragraphs.
\noindent\textbf{Compared with Megatron-1F1B.} Megatron-1F1B is deployed by 3D parallelism, including DP, EP, and PP. Models are distributed by layer chunks, and each chunk has multiple EP/DP groups. Seeing from Figure \ref{fig:overall}, \name can achieve 1.59-1.81$\times$ speedup under the model and sequence configures compared with this 3D parallelism. Interleaved PP overlaps inter-stage P2P communication but still incurs heavy intra-stage all-to-all overhead within each EP group. In contrast, \name separates attention and FFN into two stages, effectively hiding this communication cost. This design enables \name to achieve significantly higher overall throughput as discussed in Section 3.3.

\noindent\textbf{Compared with SOTA overlapping methods.} We also compare \name against other SOTA MoE overlapping methods. Firstly, we observe a consistent 1.2-1.5$\times$ speedups over Tutel \cite{tutel} and Comet \cite{zhang2025comet}. This improvement arises because both methods purely rely on operation-level overlapping, as described in Section \ref{sec:op}. In cross-node training, the computation latency of a single FFN operation accounts for only about 40–60\% of the total time of the two all-to-all communications, leaving significant unoverlapped communication overhead. By constrast, DisagMoE removes these communication bubbles by separating attention and FFN into distinct pipeline stages and overlapping token dispatch and combination across them, thereby achieving more complete module-level overlap.

Compared with DualPipe \cite{dai2024deepseekmoe}, DisagMoE delivers 1.05 - 1.13$\times$ average speedup. During warm-up and cool-down, DualPipe behaves similarly to Megatron-3D, producing large all-to-all communication bubbles along the critical path. Even in steady state, residual unoverlapped tails persist for long sequence lengths because the interleaved forward and backward passes exhibit mismatched compute-to-communication ratios between attention and FFN, limiting overlap efficiency. DisagMoE alleviates these issues by decoupling the two modules and adaptively allocating GPU and network resources to maintain balanced overlap across stages.

\begin{figure}[h]
    \centering
    \includegraphics[width=\linewidth]{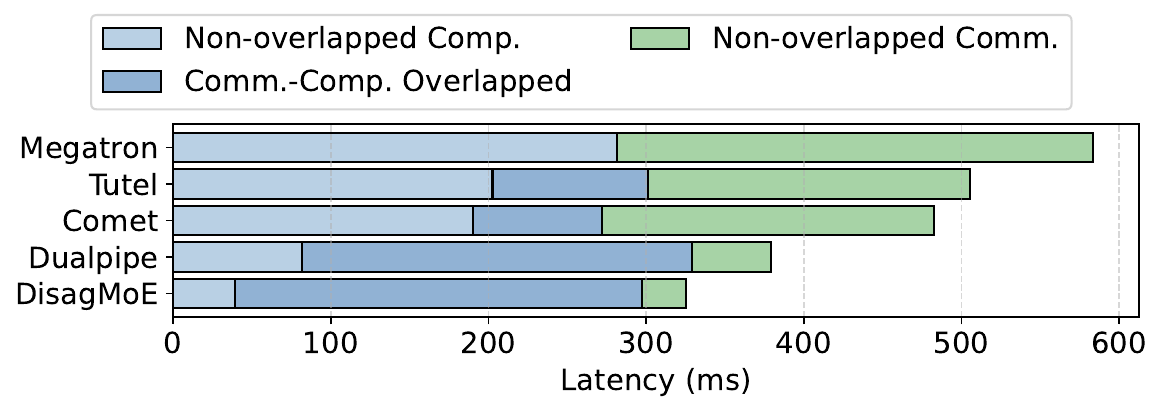}
    \caption{Latency breakdown under 8K in deepseek-moe.}
    \label{fig:expbreakdown}
\end{figure}
\noindent\textbf{Iteration time decomposition.}
Next, we analyze the contribution of each component to the total iteration time. As shown in Figure \ref{fig:expbreakdown}, \name achieves a higher level of compute-communication overlapping than baselines, reducing non-overlapped communication time by up to 88\% compared to Tutel, 75\% compared to Comet, and 45\% off compared to Dualpipe. Overall, \name expands the effective overlap region and shortens end-to-end iteration time, yielding higher overall efficiency than existing systems.

\subsection{Ablations}
\noindent\textbf{Impacts of resource allocation in AF-Pipe.} We study how the resource allocation impacts the \name's performance. Based on our Roofline Model in Sec. \ref{sec:allocation}, we change the ratio of compute resources and communication resources for attention and FFN group to search the best resource configurations for \name under different sequence length. In current GPU nodes, we allocate the compute resources by setting the \texttt{GPU\_PER\_NODES} environment variable and adjust the communication resources by changing \texttt{NCCL\_IB\_HCA} the number of NICs. Since the amount of compute and communication on each GPU depends solely on the ratio, not their absolute number. Therefore we fix the communication units number of Attention and FFN both as 16 NICs while changing the compute number of each groups.  In this allocation policy, we can dynamically adjust both of compute and communication rotios. We evaluate \name compared with Megatron baseline by the throughput per GPU.

\begin{figure}[h]
    \centering
    \includegraphics[width=\linewidth]{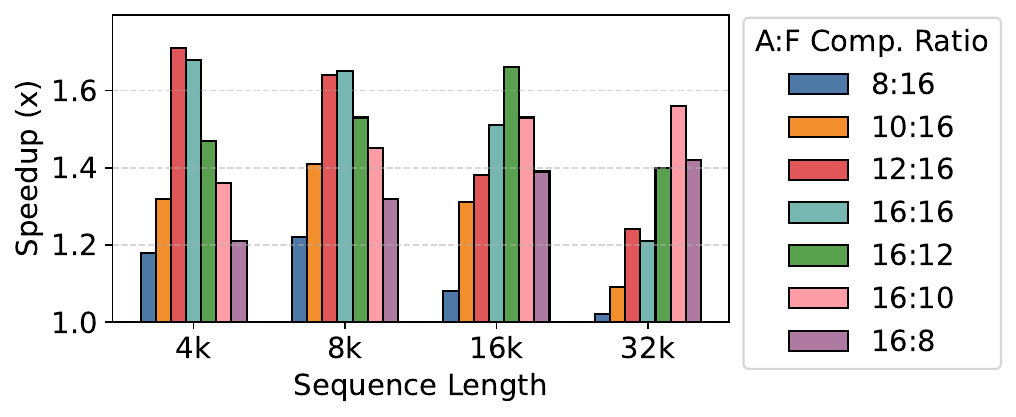}
    \caption{Effect of attention–FFN resource allocation on throughput. The y-axis shows per-GPU throughput speedups normalized over Megatron. Optimal A:F ratios vary with sequence length, where longer sequences favor more attention resources.
    }
    \label{fig:resource}
\end{figure}

As shown in Figure \ref{fig:resource}, we adjust the attention and FFN compute numbers following the resource allocator in Sec. \ref{sec:allocation}, and we observe that the optimal GPU ratio varies for different sequence lengths. In long sequence, \name’s effectiveness peaks at 16:10 for 16K input with a speed-up of 1.56x over baseline and 1.29x over the homogeneous (16:16). In this setting, we can increase the training throughput per GPU and keep the total training latency minimum. In shorter sequence, for example, 4K, we also find that since the attention and FFN computation FLOPs are closer, the optimal ratio is the balanced one (16:16) or near it. In contrast, when we forcibly enlarge or decrease the A:F ratio, the performance will be harmed since now the stages of A-Group and F-Group are showing imbalance in our AF-Pipe, which will bring extra bubbles.

\begin{figure}[h]
    \centering
    \includegraphics[width=\linewidth]{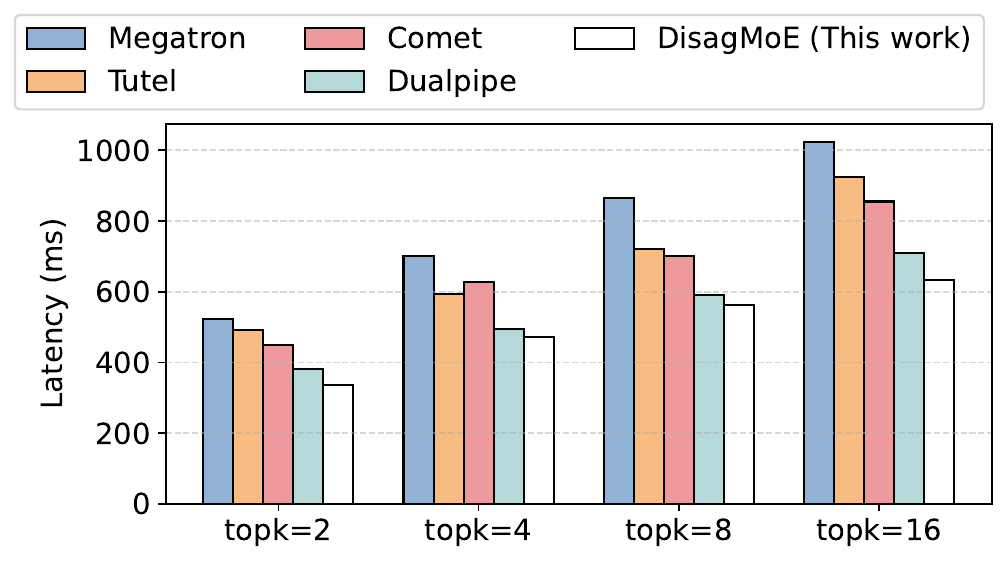}
    \caption{Impact of top-$k$ and expert-parallel (EP) size on training performance. Increasing top-$k$ raises both communication and FFN computation costs, while larger EP sizes expand cross-node communication. }
    \label{fig:exptopk}
\end{figure}

\noindent\textbf{Impacts of different top-k and EP size.}
We also study the impacts of MoE architecture configuration settings, eg, topk size and EP size. We adjust the number of EP size as well as topk to evaluate the performance of \name in total experts as 64. The results shown in Figure \ref{fig:exptopk} observe that with the topk size larger, the total training time will increase since both the communication time and the FFN computation time will increase linearly. With EP size larger, the tokens will be routed into more GPUs, increasing the communication time in meanwhile. Results show that \name consistently demonstrates better performance across different setting of topk and EP size, from 1.08-1.92x speedup compared with baselines.

\begin{figure}[t]
    \centering
    \includegraphics[width=\linewidth]{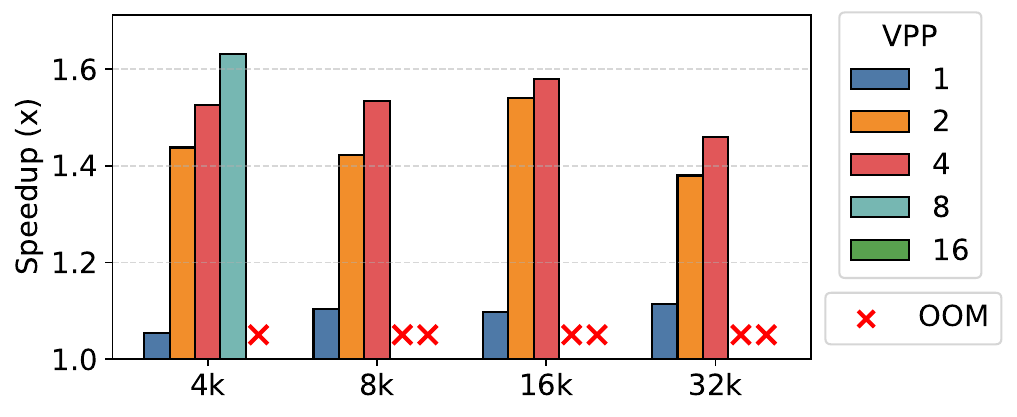}
    \caption{Effect of virtual stage size on training throughput. More virtual stages improve pipeline utilization up to the memory limit (\(v \leq 16\)), beyond which GPUs run out of memory.}
    \label{fig:expvpp}
\end{figure}

\noindent\textbf{Impacts of virtual stages in AF-Pipe.}
We further analyze how the number of interleaved virtual stages in \name affects training throughput. Each A-group and F-group holds multiple attention and FFN layers, respectively, as introduced in Section~\ref{sec:afpipe}. As shown in Fig.~\ref{fig:expvpp}, increasing the number of virtual stages enhances throughput by reducing pipeline bubbles roughly in proportion to \(v\). However, when the virtual size exceeds 16, GPU memory overflows due to the storage of parameters and activations from all 16 layers within one group. This behavior mirrors a direct extension of serving-oriented AFD systems \cite{zhu2025megascale, wang2025step} to training workloads, such as HeteroMoE \cite{wu2025hetermoe}, which quickly leads to out-of-memory (OOM) conditions.

\section{Limitations and Discussion}
\noindent\textbf{Limitations.} \name assumes pretraining-style stable shapes such as fixed sequence length and micro-batch size; supporting workloads with dynamic shapes (e.g., RL training) would require online re-allocation, which is outside the scope of this work. The current design uses a single pipeline depth $p$ shared by attention and FFN groups; asymmetric depths $(p_A, p_F)$ are a promising extension but introduce non-trivial scheduling complexity, which we leave as future work.

\noindent\textbf{Hardware-software co-design insight.} The roofline analyses in Sec.~\ref{sec:mot_roof} and Sec.~\ref{sec:allocation} surface a key observation: whether communication can be fully hidden depends on the per-component compute–communication ratio, not on raw interconnect bandwidth. Because attention and FFN exhibit a structural imbalance in this ratio, uniformly scaling bandwidth across the system cannot drive both components to the compute roof at once. This is precisely why \name disaggregates attention and FFN into separate worker groups and adaptively reallocates GPU and NIC budgets across them, so that each component is driven toward its own compute–communication ratio and the effective overlap window expands beyond what symmetric bandwidth provisioning can achieve. This view echoes the hardware proposal from DeepSeek-V4~\cite{deepseek_v4_pro}. Together, these views suggest future hardware should target favorable compute–communication ratios across components rather than scaling bandwidth alone.
\section{Conclusion}
We present \textsc{DisaggMoE}, a disaggregated MoE training system that addresses the communication bottlenecks of large-scale expert-parallel training.
By partitioning attention and FFN layers into separate worker groups and introducing AF-Pipe, a multi-stage pipeline with many-to-many communication overlap, \textsc{DisaggMoE} achieves fine-grained coordination between compute and communication. Guided by a network–compute roofline model, our adaptive allocator dynamically balances GPU and NIC resources across groups, delivering high efficiency across different sequence lengths, top-$k$, and EP sizes. Experiments on 8–16 node H800 clusters demonstrate that \textsc{DisaggMoE} consistently improves training throughput, achieving up to 1.81$\times$ speedup over Megatron-1F1B and up to 1.34$\times$ over state-of-the-art MoE overlap systems. These results highlight the effectiveness of disaggregated architecture and communication–compute co-optimization for scaling MoE training efficiently.

\clearpage

\bibliographystyle{plainnat}
\bibliography{main}

\end{document}